# SkipNet: Learning Dynamic Routing in Convolutional Networks

Xin Wang[1], Fisher Yu[1], Zi-Yi Dou[2], Trevor Darrell[1], Joseph E. Gonzalez[1]

[1] University of California, Berkeley
[2] Nanjing University

**Abstract.** While deeper convolutional networks are needed to achieve maximum accuracy in visual perception tasks, for many inputs shallower networks are sufficient. We exploit this observation by learning to skip convolutional layers on a per-input basis. We introduce SkipNet, a modified residual network, that uses a gating network to selectively skip convolutional blocks based on the activations of the previous layer. We formulate the dynamic skipping problem in the context of sequential decision making and propose a hybrid learning algorithm that combines supervised learning and reinforcement learning to address the challenges of non-differentiable skipping decisions. We show SkipNet reduces computation by $30 - 90\%$ while preserving the accuracy of the original model on four benchmark datasets and outperforms the state-of-the-art dynamic networks and static compression methods. We also qualitatively evaluate the gating policy to reveal a relationship between image scale and saliency and the number of layers skipped.

## 1 Introduction

A growing body of research in convolutional network design [10,18,28] reveals a clear trend: *deeper networks are more accurate*. Consequently, the best-performing image recognition networks have hundreds of layers and tens of millions of parameters. These very deep networks come at the expense of increased prediction cost and latency. However, a network that doubles in depth may only improve prediction accuracy by a few percentage points. While these small improvements can be critical in real-world applications, their incremental nature suggests that the majority of images do not require the doubling in network depth and that the optimal depth depends on the input image.

In this paper, we introduce SkipNets (see Fig. 1) which are modified residual networks with gating units that dynamically select which layers of a convolutional neural network should be skipped during inference. We frame the dynamic skipping problem as a sequential decision problem in which the outputs of previous layers are used to decide whether to bypass the subsequent layer. The objective in the dynamic skipping problem is then to skip as many layers as possible while retaining the accuracy of the full network. Not only can skipping policies significantly reduce the average cost of model inference they also provide insight into the diminishing return and role of individual layers.

While conceptually simple, learning an efficient skipping policy is challenging. To achieve a reduction in computation while preserving accuracy, we need to correctly bypass the unnecessary layers in the network. This inherently discrete decision is not differentiable, and therefore precludes the application of gradient based optimization.

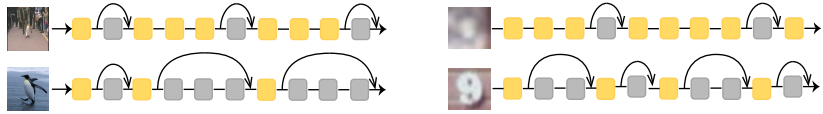

Fig. 1: The SkipNet learns to skip convolutional layers on a per-input basis. More layers are executed for challenging images (top) than easy images (bottom)

While some [2,30,31] have proposed soft approximations, we show that the subsequent hard thresholding required to reduce computation results in low accuracy.

Recent works [4,23] explored the application of reinforcement learning (RL) to learn hard decision gates. While promising, in our experiments we show that these RL based techniques are brittle, often getting stuck in poor local minima and producing networks that are not competitive with the state-of-the-art. One can also apply the reparametrization techniques [16,21], however, these approaches often find suboptimal policies partly due to the approximation error introduced by the relaxation (detailed in later sections).

We explore several SkipNet designs and introduce a hybrid learning algorithm which combines supervised learning with reinforcement learning to address the challenges of non-differentiable skipping decisions. We explicitly assign a gating module to each group of layers. The gating module maps the previous layer activations to the binary decision to skip or execute the subsequent layer. We train the gating module in two stages. First, we use a soft-max relaxation of the binary skipping decisions by adopting the reparameterization trick [16,21], and train the layers and gates jointly with standard cross entropy loss used by the original model. Then, we treat the probabilistic gate outputs as an initial skipping policy and use REINFORCE [34] to refine the policy without relaxation. In the latter stage, we jointly optimize the skipping policy and prediction error to stabilize the exploration process.

We evaluate SkipNets, using ResNets [10] as the base models, on the CIFAR-10, CIFAR-100, SVHN and ImageNet datasets. We show that, with the hybrid learning procedure, SkipNets learn skipping policies that significantly reduce model inference costs (50% on the CIFAR-10 dataset, 37% on the CIFAR-100 dataset, 86% on the SVHN dataset and 30% on the ImageNet dataset) while preserving accuracy. We compare SkipNet with several state-of-the-art models and techniques on both the CIFAR-10 and ImageNet datasets and find that SkipNet consistently outperforms the previous methods on both benchmarks. By manipulating the computational cost hyper-parameter, we show how SkipNets can be tuned for different computation constraints. Finally, we study the skipping behavior of the learned skipping policy and reveal the relation between image scale and saliency and the number of layers skipped. Our code is available at https://github.com/ucbdrive/skipnet.

## 2   Related Work

Accelerating existing convolutional networks has been a central problem in real-world deployments and several complementary approaches have been proposed. Much of this work focuses on model compression [5,9,12,20] through the application of weight sparsification, filter pruning, vector quantization, and distillation [13] to transfer knowledge to shallower networks. These methods are applied after training the initial networks and they are usually used as post-processing. Also, these optimized networks do not dynamically adjust the model complexity in response to the input. While these approaches are complimentary, we show SkipNet outperforms existing static compression techniques.

Several related efforts [6,8,29] explored dynamically scaling computation through early termination. Graves [8] explored halting in recurrent networks to save computational cost. Figurnov et al. [6] and Teerapittayanon et al. [29] proposed the use of early termination in convolutional networks. Closest to our work, Figurnov et al. [6] studied early termination in each group of blocks of ResNets. In contrast, SkipNet does not exit early but instead conditionally bypasses individual layers based on the output of the proceeding layers which we show results in a better accuracy to cost trade-off.

Another line of work [1,22,32] explores cascaded model composition. This work builds on the observation that many images can be accurately labeled with smaller models. Bolukabasi et al [1] train a termination policy for cascades of pre-trained models arranged in order of increasing costs. This standard cascaded approach fails to reuse features across classifiers and requires substantial storage overhead. Similar to the work on adaptive time computation, Bolukabasi et al. [1] also explore early termination within the network. However, in many widely used architectures (e.g., ResNet) layers are divided into groups; with some layers being more critical than others (Fig. 10a). The in-network cascading work by [1] is unable to bypass some layers while executing subsequent layers in future groups. SkipNets explore selecting layers within the network in a combinatorial way leading to a search space that is a superset of cascading.

The gating modules in SkipNets act as regulating gates for groups of layers. They are related to the gating designs in recurrent neural networks (RNN) [3,14,27]. Hochreiter et al. [14] propose to add gates to an RNN so that the network can keep important memory in network states, while Srivastava et al. [27] introduce similar techniques to convolutional networks to learn deep image representation. Both [3] and [26] apply gates to other image recognition problems. These proposed gates are "soft" in the sense that the gate outputs are continuous, while our gates are "hard" binary decisions. We show in our experiments that "hard" gating is preferable to "soft" gating for dynamic networks.

## 3  SkipNet Model Design

SkipNets are convolutional networks in which individual layers are selectively included or excluded for a given input. The per-input selection of layers is accomplished using small gating networks that are interposed between layers. The gating networks map the output of the previous layer or group of layers to a binary decision to execute or bypass the subsequent layer or group of layers as illustrated in Fig. 2.

More precisely, let $\mathbf{x}^i$ be the input and $F^i(\mathbf{x}^i)$ be the output of the $i^{\text{th}}$ layer or group of layers, then we define the output of the gated layer (or group of layers) as:

$$\mathbf{x}^{i+1} = G^i(\mathbf{x}^i)F^i(\mathbf{x}^i) + (1 - G^i(\mathbf{x}^i))\mathbf{x}^i, \tag{1}$$

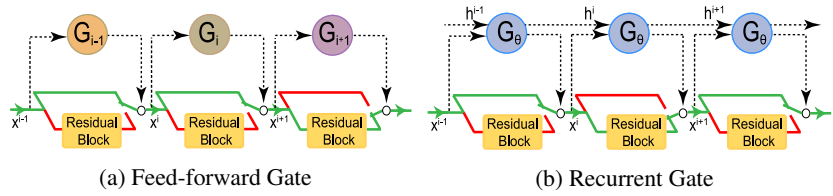

(a) Feed-forward Gate        (b) Recurrent Gate

Fig. 2: We study SkipNets with two gating designs. (a) Each residual block has a unique gating module and gate parameters. (b) A unified recurrent gate is shared across blocks

where $G^i(\mathbf{x}^i) \in \{0, 1\}$ is the gating function for layer $i$. In order for Eq. 1 to be well defined, we require $F^i(\mathbf{x}^i)$ and $\mathbf{x}^i$ to have the same dimensions. This requirement is satisfied by commonly used residual network architectures where

$$\mathbf{x}^{i+1}_{\text{ResNet}} = F^i(\mathbf{x}^i_{\text{ResNet}}) + \mathbf{x}^i_{\text{ResNet}}, \qquad (2)$$

and can be addressed by pooling $\mathbf{x}^i$ to match the dimensions of $F^i(\mathbf{x}^i)$.

The gating network design needs to be both sufficiently expressive to accurately determine which layers to skip while also being computationally cheap. To address this trade-off between accuracy and computational cost we explore a range of gating network designs (Sec. 3.1) spanning feed-forward convolutional architectures to recurrent networks with varying degrees of parameter sharing. In either case, estimating the gating network parameters is complicated by the discrete gate decisions and the competing goals of maximizing accuracy and minimizing cost. To learn the gating network we introduce a two stage training algorithm that combines supervised pre-training (Sec. 3.3) with based policy optimization (Sec. 3.2) using a hybrid reward function that combines prediction accuracy with the computational cost.

### 3.1   Gating Network Design

In this paper, we evaluate two feed-forward convolutional gate designs (Fig. 2a). The *FFGate-I* (Fig. 11a) design is composed of two $3 \times 3$ convolutional layers with stride of 1 and 2 respectively followed by a global average pooling layer and a fully connected layer to output a single dimension vector. To reduce the gate computation, we add a $2 \times 2$ max pooling layer prior to the first convolutional layer. The overall computational cost of FFGate-I is roughly $19\%$ of the residual blocks [10] used in this paper. As a computationally cheaper alternative, we also introduce the *FFGate-II* (Fig. 11b), consisting of one $3 \times 3$ stride 2 convolutional layer followed by the same global average pooling and fully connected layers as *FFGate-I*. The computational cost of *FFGate-II* is $12.5\%$ of the cost of the residual block. In our experiments, we use *FFGate-II* for networks with more than 100 layers and *FFGate-I* for shallower networks.

The feed-forward gate design is still relatively costly to compute and does not leverage the decisions from previous gates. Therefore, we introduce a *recurrent gate* (RNNGate) design (Fig. 11c) which enables parameter sharing and allows gates to re-use computation across stages. We first apply global average pooling on the input feature map of the gates and then linearly project the feature to the input size of 10. We adopt a single layer *Long Short Term Memory* [14] (LSTM) with hidden unit size of 10. At each gate, we project the LSTM output to a one-dimensional vector to compute the final gate decision. Compared to the cost of computing residual blocks, the cost of this recurrent gate design is negligible (roughly 0.04% of the computation of residual blocks).

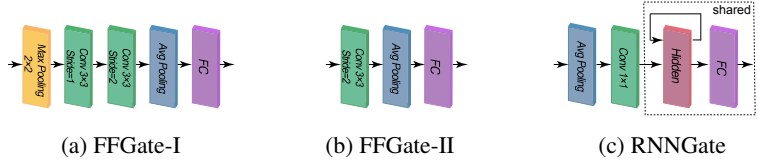

(a) FFGate-I  (b) FFGate-II  (c) RNNGate

Fig. 3: Gate designs. (a) FFGate-I contains two 3x3 convolutional layers and has roughly 19% of the computation of the residual block. (b) FFGate-II is composed of one convolutional layer with stride of 2 and has about 12.5% of the computation of residual blocks. (c) RNNGate contains a one-layer LSTM with both input and hidden unit size of 10. The cost of the RNNGate is negligible at 0.04% of the cost of the residual blocks.

In our later experiments, we find that the recurrent gate dominates the feed-forward gates in both prediction accuracy and computation cost. We also evaluated simpler feed-forward gate designs without convolution layers and while these matched the computation cost of the recurrent gates the prediction accuracy suffered. We conjecture that the recurrent gate design better captures the cross-layer dependencies.

### 3.2 Skipping Policy Learning with Hybrid RL

During inference the most likely action is taken from the probability distribution encoded by each gate: *the layer is skipped or executed*. This inherently *discrete* and therefore non-differentiable decision process creates unique challenges for how we train SkipNets. A natural approximation, similar to that used in Highway Networks [27], would be to use differentiable soft-max decisions during training and then revert to hard decisions during inference. While this approach enables gradient based training, it results in poor prediction accuracy (Sec. 4.3) as the network parameters are not optimized for the subsequent hard-gating during inference. We therefore explore the use of reinforcement learning to learn the model parameters for the non-differentiable decision process.

Because SkipNets make a sequence of discrete decisions, one at each gated layer, we frame the task of estimating the gating function in the context of policy optimization through reinforcement learning. We define the skipping policy:

$$\pi(\mathbf{x}^i, i) = \mathbb{P}(G^i(\mathbf{x}^i) = g_i) \tag{3}$$

as a function from the input $\mathbf{x}^i$ to the probability distribution over the gate action $g_i$ to execute ($g_i = 1$) or skip ($g_i = 0$) layer $i$. We define a sample sequence of gating decisions drawn from the skipping policy starting with input $\mathbf{x}$ as:

$$\mathbf{g} = [g_1, \ldots, g_N] \sim \pi_{F_\theta}, \tag{4}$$

where $F_\theta = \left[F_\theta^1, \ldots, F_\theta^N\right]$ is the sequence of network layers (including the gating modules) parameterized by $\theta$ and $\mathbf{g} \in \{0,1\}^N$. The overall objective is defined as

$$\begin{aligned} \min \mathcal{J}(\theta) &= \min \mathbb{E}_\mathbf{x} \mathbb{E}_\mathbf{g} L_\theta(\mathbf{g}, \mathbf{x}) \\ &= \min \mathbb{E}_\mathbf{x} \mathbb{E}_\mathbf{g} \left[\mathcal{L}(\hat{y}(\mathbf{x}, F_\theta, \mathbf{g}), y) - \frac{\alpha}{N} \sum_{i=1}^N R_i\right], \end{aligned} \quad (5)$$

where $R_i = (1 - g_i)C_i$ is the reward of each gating module. The constant $C_i$ is the cost of executing $F^i$ and the term $(1 - g_i)C_i$ reflects the reward associated with *skipping* $F^i$. In our experiments, all $F^i$ have the same cost and so we set $C_i = 1$. Finally $\alpha$ is a tuning parameter that allows us to trade-off the competing goals of minimizing the prediction loss and maximizing the gate rewards.

To optimize this objective, we can derive the gradients with respect to $\theta$ as follows. We define $\pi_{F_\theta}(\mathbf{x}) = p_\theta(\mathbf{g}|\mathbf{x})$, $\mathcal{L} = \mathcal{L}(\hat{y}(\mathbf{x}, F_\theta, \mathbf{g}), y)$ and $r_i = -[\mathcal{L} - \frac{\alpha}{N}\sum_{j=i}^N R_j]$.

$$\begin{aligned} \nabla_\theta \mathcal{J}(\theta) &= \mathbb{E}_\mathbf{x} \nabla_\theta \sum_\mathbf{g} p_\theta(\mathbf{g}|\mathbf{x}) L_\theta(\mathbf{g}, \mathbf{x}) \\ &= \mathbb{E}_\mathbf{x} \sum_\mathbf{g} p_\theta(\mathbf{g}|\mathbf{x}) \nabla_\theta \mathcal{L} + \mathbb{E}_\mathbf{x} \sum_\mathbf{g} p_\theta(\mathbf{g}|\mathbf{x}) \nabla_\theta \log p_\theta(\mathbf{g}|\mathbf{x}) L_\theta(\mathbf{g}, \mathbf{x}) \\ &= \mathbb{E}_\mathbf{x} \mathbb{E}_\mathbf{g} \nabla_\theta \mathcal{L} - \mathbb{E}_\mathbf{x} \mathbb{E}_\mathbf{g} \sum_{i=1}^N \nabla_\theta \log p_\theta(g_i|\mathbf{x}) r_i. \end{aligned} \quad (6)$$

The first part of Eq. 6 corresponds to the supervised learning loss while the second part corresponds to the REINFORCE [34] gradient where $r_i$ is the cumulative future rewards associated the gating modules. We refer to this combined reinforcement learning and supervised learning procedure as *hybrid reinforcement learning*. In practice, we may relax the reward $\hat{r}_i = -\left[\beta\mathcal{L} - \frac{\alpha}{N}\sum_{j=i}^N R_j\right]$ to scale down the influence of the prediction loss as this hybrid reinforcement learning is followed by the supervised pre-training that will be discussed in the next section. We set $\beta = \frac{\alpha}{N}$ in our experiments on ImageNet and $\beta = 1$ for other datasets.

### 3.3   Supervised Pre-training

Optimizing Eq. 8 starting from random parameters also consistently produces models with poor prediction accuracy (Sec. 4.3). We conjecture that the reduced ability to learn is due to the interaction between policy learning and image representation learning. The gating policy can over-fit to early features limiting future feature learning.

To provide an effective supervised initialization procedure we introduce a form of supervised pre-training that combines hard-gating during the forward pass with *soft-gating* during backpropagation. We relax the gate outputs $G(\mathbf{x})$ in Eq. 1 to continuous values (i.e. approximating $G(\mathbf{x})$ by $S(\mathbf{x}) \in [0, 1]$). We round the output gating probability of the skipping modules in the forward pass. During backpropagation we use the soft-max approximation [16,21] and compute the gradients with respect to soft-max outputs. The

---

**Algorithm 1:** Hybrid Learning Algorithm (HRL+SP)

**Input:** A set of images **x** and labels **y**
**Output:** Trained SkipNet
1. Supervised pre-training (Sec. 3.3)
   $\theta_{SP} \leftarrow \text{SGD}(L_{\text{Cross-Entropy}}, \text{SkipNet-}G_{\text{relax}}(\mathbf{x}))$
2. Hybrid reinforcement learning (Sec. 3.2)
   Initialize $\theta_{HRL+SP}$ with $\theta_{SP}$
   $\theta_{HRL+SP} \leftarrow \text{REINFORCE}(\mathcal{J}, \text{SkipNet-}G(\mathbf{x}))$

---

relaxation procedure is summarized by:

$$G_{\text{relax}}(\mathbf{x}) = \begin{cases} \mathbb{I}(S(\mathbf{x}) \geq 0.5), & \text{forward pass} \\ S(\mathbf{x}), & \text{backward pass} \end{cases}, \quad (7)$$

where $\mathbb{I}(\cdot)$ is the indicator function. This hybrid form of supervised pre-training is able to effectively leverage labeled data to initialize model parameters for both the base network and the gating networks. After supervised pre-training we then apply the REINFORCE algorithm to refine the model and gate parameters improving prediction accuracy and further reducing prediction cost. Our two stage hybrid algorithm is given in Alg. 1.

## 4 Experiments

We evaluate a range of SkipNet architectures and our proposed training procedure on four image classification benchmarks: CIFAR-10/100 [17], SVHN [24] and ImageNet 2012 [25]. We construct SkipNets from ResNet models [10] by introducing hard gates between residual blocks. In Sec. 4.1, we evaluate the performance of SkipNets with both gate designs and compare SkipNets with the state-of-the-art models including dynamic networks and static compression networks which are also complementary approaches to our methods. We also compare our approach with baselines inspired by [15] to demonstrate the effectiveness of the learned skipping policy. In Sec. 4.2, we decipher the dynamic essence of SkipNets with extensive qualitative study and analysis to reveal the relation between image scale and saliency and number of layers skipped. In Sec. 4.3, we discuss the effectiveness of the proposed learning algorithm and gating designs.

**Datasets:** Tab. 1 summarizes the statistics of datasets used in this paper. We follow the common data augmentation scheme (mirroring/shifting) that is adopted for CIFAR and ImageNet datasets [7,19,33]. For the SVHN dataset, we use both the training and provided extra dataset for training and did not perform data augmentation [15]. For preprocessing, we normalize the data with the channel means and standard deviations.

---

[3] 531,131 of the images are extra images of SVHN for additional training.

| Table 1: Dataset statistics | | | |
|---|---|---|---|
| Dataset | # Train | # Test | # Classes |
| CIFAR-10 | 50k | 10k | 10 |
| CIFAR-100 | 50k | 10k | 100 |
| SVHN | 604k[3] | 26k | 10 |
| ImageNet | 1.28m | 50k | 1k |

| Table 2: Top 1 accuracy of ResNets (R for short) | | | | | |
|---|---|---|---|---|---|
| Model | CIFAR-10 | CIFAR-100 | SVHN | Model | ImageNet |
| R-38 | 92.50% | 68.54% | 97.94% | R-34 | 73.30% |
| R-74 | 92.95% | 70.64% | 97.92% | R-50 | 76.15% |
| R-110 | 93.60% | 71.21% | 98.09% | R-101 | 77.37% |
| R-152 | - | - | 98.14% | - | - |

**Models:** For CIFAR and SVHN, we use the ResNet [10] architecture with $6n+2$ stacked weighted layers for our base models and choose $n =$ {6, 12, 18, 25} to construct network instances with depth of {38, 74, 110, 152}. For ImageNet, we evaluate ResNet-34, ResNet-50 and ResNet-101 as described in [10]. We denote our model at depth $x$ by SkipNet-$x$. In addition, we add +SP and +HRL to indicate whether supervised pre-training or hybrid reinforcement learning were used. If no modifier is provided then we conduct the full two stage training procedure. Finally we will also use +FFGate and +RNNGate to indicate which gating design is being used. If not specified, RNNGate is used. We summarize the accuracy of the base models in Tab. 2. In later sections, we demonstrate SkipNets can preserve the same accuracy (within a variance of 0.5%).

**Training:** Our two-stage training procedure combines supervised pre-training and policy refinement with hybrid reinforcement learning. In the first stage, we adopt the same hyper-parameters used in [10] for CIFAR and ImageNet and [15] for SVHN.

For the policy refining stage, we use the trained models as initialization and optimize them with the same optimizer with decreased learning rate of 0.0001 for all datasets. We train a fixed number of iterations (10k iterations for the CIFAR datasets, 50 epochs for the SVHN dataset and 40 epochs for the ImageNet dataset) and report the test accuracy evaluated at termination. The training time of the supervised pre-training stage is roughly the same as training the original models without gating. Our overall training time is slightly longer with an increase of about 30-40%.

### 4.1   SkipNet Performance Evaluation

In this subsection, we first provide the overall computation reduction of SkipNets on four benchmark datasets to demonstrate SkipNet achieves the primary goal of reducing computation while preserving full network prediction accuracy. We also show that by adjusting $\alpha$, SkipNet can meet different computational cost and accuracy requirements. For horizontal comparison, we show SkipNet outperforms a set of state-of-the-art dynamic network and static compression techniques on both ImageNet and CIFAR-10.

**Computation reduction while preserving full network accuracy:** Fig. 4 and Fig. 5a show the computation cost (including the computation of the gate networks), measured in *floating point operations* (FLOPs), of the original ResNets and SkipNets with both feed-forward and recurrent gate designs with $\alpha$ tuned to match the same accuracy (variance less than 0.5%). The trade-off between accuracy and computational cost will be discussed later. Following [10], we only consider the multiply-adds associated with convolution operations as others have negligible impact on cost.

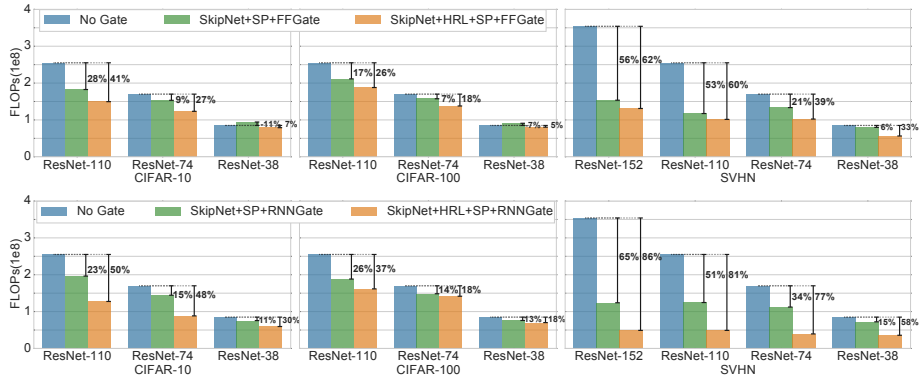

Fig. 4: Computation reduction of SkipNet+SP and SkipNet+HRL+SP with feed-forward gates and recurrent gates while preserving the full network accuracy. The computation cost includes the computation of gates. We are able to reduce computation costs by 50%, 37% and 86% of the deepest models on the CIFAR-10, 100 and SVHN data. Compared to using SP only, fine-tuning with HRL can gain another 10% or more computation reduction. Since feed-forward gates are more expensive, SkipNets with recurrent gates generally achieve greater cost savings

We observe that the hybrid reinforcement learning (HRL) with supervised pre-training (SkipNet+HRL+SP) is able to substantially reduce the cost of computation. Overall, for the deepest model on each dataset, SkipNet-110+HRL+SP with recurrent gates reduces computation on the CIFAR-10 and CIFAR-100 datasets by 50% and 37% respectively. The largest SkipNet-152+HRL+SP model with recurrent gates reduces computation on the SVHN dataset by 86%. On the ImageNet data, the SkipNet-101+HRL+SP using recurrent gates is able to reduce computation by 30%. Interestingly, as noted earlier, even in the absence of the cost regularization in the objective, the supervised pre-training of the SkipNet architecture consistently results in reduced prediction costs. One way to explain it is that the shallower network is easier to train and thus more favorable. We also observe that deeper networks tend to experience greater cost reductions which supports our conjecture that only a small fraction of inputs require extremely deep networks.

**Trade-off computational cost and accuracy:** Eq. 8 introduces the hyper-parameter $\alpha$ to balance the computational cost and classification accuracy. In Fig. 5b we plot the accuracy against the average number of skipped layers for different values of $\alpha$ from 0.0 to 4.0 on ImageNet. We observe similar patterns on other datasets and details can be found in the supplementary material. By adjusting $\alpha$, one can trade-off computation and accuracy to meet various computation or accuracy requirements.

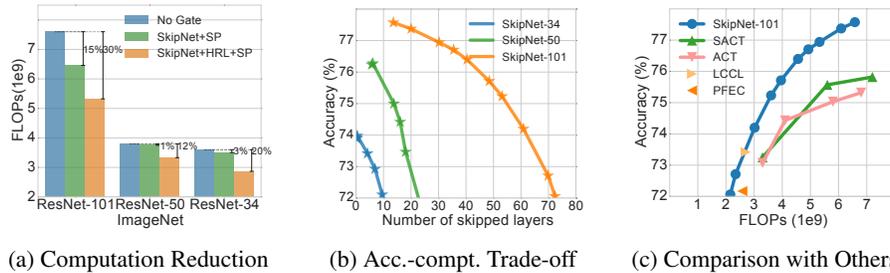

(a) Computation Reduction  (b) Acc.-compt. Trade-off  (c) Comparison with Others

Fig. 5: ImageNet evaluation. (a) Computation reduction (12 - 30%) achieved by SkipNets with RNNGates while preserving full network accuracy. (b) Trade-off between accuracy and cost under different $\alpha$. With small $\alpha$, the computation drops faster than the decrease of accuracy. (c) Comparison of SkipNet with state-of-the-art models. SkipNet consistently outperforms existing approaches on both benchmarks under various trade-off between computational cost and prediction accuracy

**Comparison with state-of-the-art models:** We compare SkipNet with existing state-of-the-art models on both ImageNet (Fig. 5c) and CIFAR-10 (Fig. 6c). The SACT and ACT models proposed by [6] are adaptive computation time models that attempt to terminate computation early in each group of blocks of ResNets (Sec. 2). In addition, we compare SkipNet with static compression techniques: PFEC [20] and LCCL [5] which are also complementary approaches to our method.

As shown in Fig. 5c, SkipNet-101 outperforms SACT and ACT models by a large margin on the ImageNet benchmark even though they are using the recent more accurate pre-activation [11] ResNet-101 as the base model. We hypothesize that increased flexibility afforded by the skipping model formulation enables the SkipNet design to outperform SACT and ACT. Similar patterns can be observed on CIFAR-10 in Fig. 6c.[4]

For comparison with the static compression techniques, we plot the computation FLOPs and the accuracy of the compressed residual networks (may have different depths from what we used in this paper) in Fig. 5. Though the static compression techniques are complementary approaches, SkipNet performs similar to or better than these techniques. Note that, though LCCL [5] uses shallower and cheaper ResNets (34 layers on ImageNet and 20, 32, 44 layers on CIFAR-10), our approach still obtains comparable performance.

**Comparison with stochastic depth network variant:** Huang et al.[15] propose stochastic depth networks which randomly drop layers for a each training mini-batch and revert to using the full network for inference. The original goal of the stochastic depth model is to avoid gradient vanishing and speed up training. A natural variant of this model in order to reduce inference computation cost is to skip blocks randomly with a chosen ratio in both training and inference phases referred as *SDV*. We compare SkipNet to SDV on both the CIFAR-10 and CIFAR-100 datasets shown in Fig. 6a and 6b. SkipNet outperforms SDV by a large margin under networks with different depths.

---
[4] We obtain the CIFAR-10 results by running the code provided by the authors

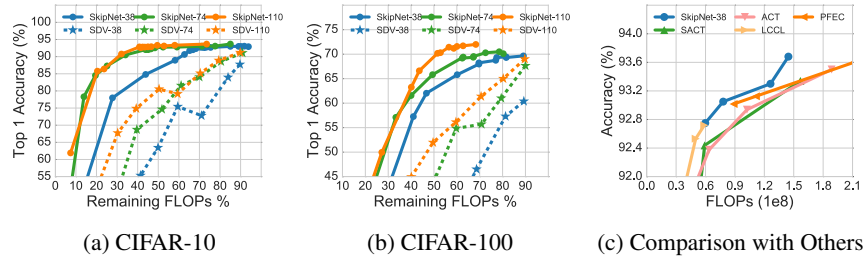

(a) CIFAR-10  (b) CIFAR-100  (c) Comparison with Others

Fig. 6: Comparison on CIFAR. (a) Comparison on CIFAR-10 with a variant of the stochastic depth model (SDV) that randomly drops blocks with chosen ratios during training and testing. The learned policy of SkipNet outperforms the baseline under various skipping ratios (b) Comparison on CIFAR-100 with SDV (c) Comparison of SkipNet with the state-of-the-art models on CIFAR-10. SkipNet is consistently matches or out-performs state-of-the-art models

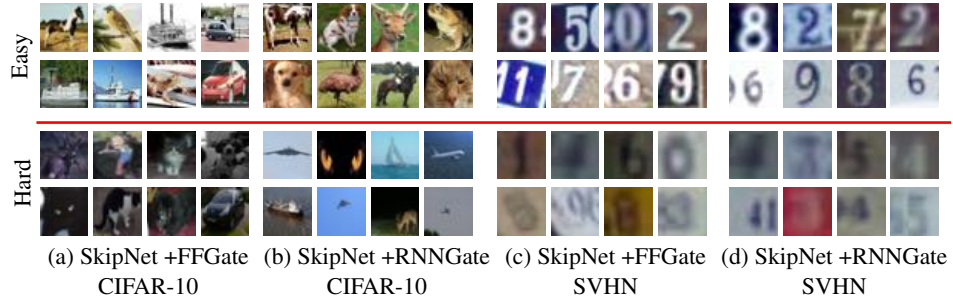

(a) SkipNet +FFGate CIFAR-10   (b) SkipNet +RNNGate CIFAR-10   (c) SkipNet +FFGate SVHN   (d) SkipNet +RNNGate SVHN

Fig. 7: Visualization of *easy* and *hard* images in the CIFAR-10 and SVHN with SkipNet-74. The top two rows are easy examples (with more than 15 layers skipped) and the bottom two rows are hard examples (with fewer than 8 layers skipped). Easy examples are brighter and clearer while hard examples tend to be dark and blurry

### 4.2 Skipping Behavior Analysis and Visualization

In this subsection, we investigate the key factors associated with the dynamic skipping and qualitatively visualize their behavior. We study the correlation between block skipping and the input images in the following aspects: (1) qualitative difference between images (2) the scale of the inputs and (3) prediction accuracy per category. We find that SkipNet skips more aggressively on inputs with smaller scales and on brighter and clearer images. Moreover, more blocks are skipped for classes with high accuracy.

**Qualitative difference between inputs:** To better understand the learned skipping patterns, we cluster the images that SkipNets skip many layers (treated as *easy* examples) and keep many layers (treated as *hard* examples) in Fig. 7 for both CIFAR-10 and SVHN. Interestingly, we find that images within each cluster share similar characteristics with respect to saliency and clarity. On both datasets, we observe that the easy examples are

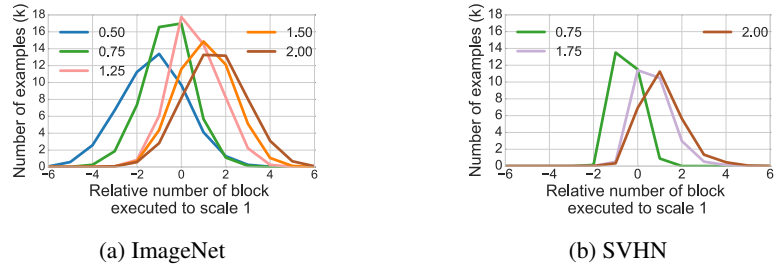

(a) ImageNet    (b) SVHN

Fig. 8: Distribution of number of blocks executed with multi-scale inputs. The $x$-axis is the relative number of blocks executed to scale 1 (#block kept at scale $s$ - #block kept at scale 1). More blocks are executed for inputs with larger scales.

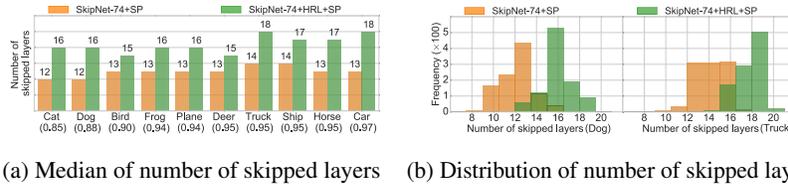

(a) Median of number of skipped layers    (b) Distribution of number of skipped layers

Fig. 9: Correlation of number of skipped layers and the level of complexity of different classes. (a) SkipNets skip more layers on classes with higher accuracy. (b) The distribution of the *hard* class (dogs) is left-skewed while the *easy* class (trucks) is right-skewed

more salient (bright, clear and with high contrast) while the hard examples are dark and blurry which are even hard for humans to recognize. These findings suggest that SkipNet can identify the visual difference of the inputs and skip layers accordingly.

**Input scales:** We conjecture the input scale affects the skipping decisions of the gates. To verify this hypothesis, we conduct multi-scale testing of trained models on the ImageNet and SVHN datasets. We plot the distribution of the number of blocks executed of different input scales relative to the original scale 1 used in other experiments. We observe on both datasets that the distributions of smaller scales are skewed left (executing less blocks than the model with input scale 1) while the distributions of larger scales are skewed right (more block executed). This observation matches the intuition that inputs with larger scale require larger receptive field and thus need to execute more blocks. Another interpretation is that SkipNet dynamically selects layers with appropriate receptive field sizes for the given inputs with different input scales.

**Prediction accuracy per category:** We further study the correlation of skipping behaviors and the prediction accuracy per class on CIFAR-10. The conjecture is that the SkipNet skips more on easy classes (class with high accuracy, e.g., truck class) while skipping less on hard classes (class with low accuracy, e.g., cat and dog classes). We plot the median of number of skipped layers in each class for SkipNet+SP and SkipNet+HRL+SP in Fig. 9a. It shows that while all classes tend to skip more aggressively after applying HRL, SkipNets tend to skip more layers on easy classes. Fig. 9b indicates that the distribution of hard classes (e.g. dog class) are skewed left, whereas easier classes (e.g. truck class) are skewed right as SkipNet tends to skip more layers on easier classes.

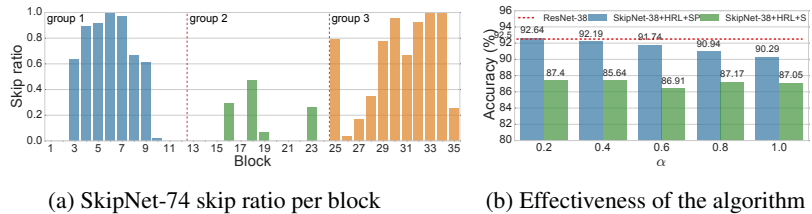

(a) SkipNet-74 skip ratio per block    (b) Effectiveness of the algorithm

Fig. 10: Visualization and analysis on CIFAR-10. (a) Visualization of the skipping ratio per block of SkipNet-74. The first and last few blocks are more critical in each group; group 2 is heavily utilized. (b) Accuracy of SkipNet-38+HRL+SP, SkipNet-38+HRL+S and SkipNet-38+RL with FFGates for different $\alpha$ values. SkipNet-38+RL (not plotted) can only achieve $\sim 10\%$ accuracy which indicates with pure RL, SkipNet fails to learn the correct feature representations. Moreover, the accuracy improves consistently with SP compared with random initialization.

**Skip ratio of different blocks:** We visualize the skip ratio of different blocks in SkipNet in Fig. 10a on CIFAR-10. Visualizations of SkipNets on other datasets (e.g. ImageNet, CIFAR-100) can be found in the supplementary material. The ResNet model can be divided into 3 groups where blocks in the same group have the same feature map size and tend to have similar functionality. Interestingly, we observe less skipping in group 2 than in groups 1 and 3 suggesting group 2 may be more critical to feature extraction.

### 4.3 SkipNet Design and Algorithm Analysis

**Effectiveness of hybrid learning algorithm:** We compare the performances of SkipNet-38 trained using basic RL, hybrid RL from scratch (HRL+S), and hybrid RL plus supervised pre-training (HRL+SP) on CIFAR-10 in Fig. 10b. For SkipNet+HRL+S and SkipNet+RL, we train both networks for 80k iterations to match the total training steps of the two-stage training of SkipNet+HRL+SP.

First, we were unable to train the model using the pure RL approach (SkipNet-38+RL accuracy was roughly $10\%$). This provides strong evidence for the importance of supervision in complex vision tasks. Second, SkipNet-38+HRL+SP consistently achieves higher accuracy than SkipNet-38+HRL+S. The accuracy of SkipNet-38+HRL+S is lower than the accuracy of the original ResNet-38 model even with very small $\alpha$. This suggests that supervised pre-training can provide a more effective initialization which helps the HRL stage to focus more on skipping policy learning.

Table 3: "Hard" gating *vs* "soft" gating. With similar computation, SkipNet (S for short) with "hard" gating has much better accuracy than SkipNet with "soft" gating.

| Data | Model | Acc. (%) | FLOPs (1e8) | Data | Model | Acc.(%) | FLOPs (1e8) |
|---|---|---|---|---|---|---|---|
| CIFAR-10 | S-38-Hd | 90.83 | 0.58 | CIFAR-100 | S-38-Hd | 67.68 | 0.50 |
| | S-38-St | 66.67 | 0.61 | | S-38-St | 21.70 | 0.62 |
| | S-74-Hd | 92.38 | 0.92 | | S-74-Hd | 67.79 | 0.61 |
| | S-74-St | 52.29 | 1.03 | | S-74-St | 25.47 | 0.89 |
| | S-110-Hd | 88.11 | 0.18 | | S-110-Hd | 63.66 | 0.96 |
| | S-110-St | 23.44 | 0.05 | | S-110-St | 9.84 | 1.00 |

**"Hard" gating and "Soft" gating design** During supervised pre-training, we can either treat gate outputs as "hard" (Sec. 3.3) or "soft" (Sec. 3.2).For "soft" gating, continuous gating probabilities are adopted for training but discretized values are used for inference to achieve the desired computation reduction. In Tab. 3, we show the classification accuracy of SkipNet with "hard" (SkipNet-Hd) and "soft" gating (SkipNet-St) under similar computation cost[5]. SkipNet-Hd achieves much higher accuracy than SkipNet-St which may be due to the inconsistency between training and inference with soft gating.

## 5   Conclusion

We introduced SkipNet architecture that learns to dynamically skip redundant layers on a per-input basis, without sacrificing prediction accuracy. We framed the dynamic execution problem as a sequential decision problem. To address the inherent non-differentiability of dynamic execution, we proposed a novel hybrid learning algorithm which combines the strengths of supervised and reinforcement learning.

We evaluated the proposed approach on four benchmark datasets, showing that Skip-Nets reduce computation substantially while preserving the original accuracy. Compared to both state-of-the-art dynamic models and static compression techniques, SkipNets obtain better accuracy with lower computation. Moreover, we conducted a range of ablation study to further evaluate the proposed network architecture and algorithm.

The dynamic architectures offer the potential to be more computationally efficient and improve accuracy by specializing and reusing individual components. We believe that further study in this area will be critical to the long term progress in machine learning and computer vision.

## Acknowledgements

We would like to thank the ECCV reviewers for their excellent feedback. This research was funded by the NSF CISE Expeditions Award CCF-1730628 and generous gifts from Alibaba, Amazon Web Services, Ant Financial, Arm, CapitalOne, Ericsson, Facebook, Google, Huawei, Intel, Microsoft, Scotiabank, Splunk and VMware.

---

[5] We tune SkipNet-Hd to match the computation of SkipNet-St.

# References

1. Bolukbasi, T., Wang, J., Dekel, O., Saligrama, V.: Adaptive neural networks for efficient inference. In: Proceedings of the 34th International Conference on Machine Learning. pp. 527–536 (2017) 3
2. Chan, W., Jaitly, N., Le, Q., Vinyals, O.: Listen, attend and spell: A neural network for large vocabulary conversational speech recognition. In: Acoustics, Speech and Signal Processing (ICASSP), 2016 IEEE International Conference on. pp. 4960–4964. IEEE (2016) 2
3. Dauphin, Y.N., Fan, A., Auli, M., Grangier, D.: Language modeling with gated convolutional networks. In: International Conference on Machine Learning. pp. 933–941 (2017) 3
4. Dhingra, B., Li, L., Li, X., Gao, J., Chen, Y.N., Ahmed, F., Deng, L.: Towards end-to-end reinforcement learning of dialogue agents for information access. In: Proceedings of the 55th Annual Meeting of the Association for Computational Linguistics. vol. 1, pp. 484–495 (2017) 2
5. Dong, X., Huang, J., Yang, Y., Yan, S.: More is less: A more complicated network with less inference complexity. In: Proceedings of the IEEE Conference on Computer Vision and Pattern Recognition. pp. 5840–5848 (2017) 3, 10
6. Figurnov, M., Collins, M.D., Zhu, Y., Zhang, L., Huang, J., Vetrov, D., Salakhutdinov, R.: Spatially adaptive computation time for residual networks. In: The IEEE Conference on Computer Vision and Pattern Recognition (July 2017) 3, 10
7. Goodfellow, I.J., Warde-Farley, D., Mirza, M., Courville, A., Bengio, Y.: Maxout networks. In: Proceedings of the 30th International Conference on Machine Learning. pp. III–1319 (2013) 7
8. Graves, A.: Adaptive computation time for recurrent neural networks. NIPS 2016 Deep Learning Symposium (2016) 3
9. Han, S., Mao, H., Dally, W.J.: Deep compression: Compressing deep neural networks with pruning, trained quantization and huffman coding. International Conference on Learning Representations (2016) 3
10. He, K., Zhang, X., Ren, S., Sun, J.: Deep residual learning for image recognition. In: Proceedings of the IEEE conference on computer vision and pattern recognition. pp. 770–778 (2016) 1, 2, 4, 7, 8, 9
11. He, K., Zhang, X., Ren, S., Sun, J.: Identity mappings in deep residual networks. In: European Conference on Computer Vision. pp. 630–645 (2016) 10
12. He, Y., Zhang, X., Sun, J.: Channel pruning for accelerating very deep neural networks. In: International Conference on Computer Vision (ICCV). vol. 2, p. 6 (2017) 3
13. Hinton, G., Vinyals, O., Dean, J.: Distilling the knowledge in a neural network. NIPS 2014 Deep Learning Workshop (2014) 3
14. Hochreiter, S., Schmidhuber, J.: Long short-term memory. Neural computation **9**(8), 1735–1780 (1997) 3, 5
15. Huang, G., Sun, Y., Liu, Z., Sedra, D., Weinberger, K.Q.: Deep networks with stochastic depth. In: European Conference on Computer Vision. pp. 646–661 (2016) 7, 8, 10
16. Jang, E., Gu, S., Poole, B.: Categorical reparameterization with gumbel-softmax. International Conference on Learning Representations (2017) 2, 6
17. Krizhevsky, A.: Learning multiple layers of features from tiny images. Tech. rep. (2009) 7
18. Krizhevsky, A., Sutskever, I., Hinton, G.E.: Imagenet classification with deep convolutional neural networks. In: Advances in neural information processing systems. pp. 1097–1105 (2012) 1
19. Lee, C.Y., Gallagher, P.W., Tu, Z.: Generalizing pooling functions in convolutional neural networks: Mixed, gated, and tree. In: Artificial Intelligence and Statistics. pp. 464–472 (2016) 7

20. Li, H., Kadav, A., Durdanovic, I., Samet, H., Graf, H.P.: Pruning filters for efficient convnets. International Conference on Learning Representations (2017) 3, 10
21. Maddison, C.J., Mnih, A., Teh, Y.W.: The concrete distribution: A continuous relaxation of discrete random variables. International Conference on Learning Representations (2017) 2, 6
22. McGill, M., Perona, P.: Deciding how to decide: Dynamic routing in artificial neural networks. In: Proceedings of the 34th International Conference on Machine Learning. pp. 2363–2372 (2017) 3
23. Mnih, V., Heess, N., Graves, A., et al.: Recurrent models of visual attention. In: Advances in neural information processing systems. pp. 2204–2212 (2014) 2
24. Netzer, Y., Wang, T., Coates, A., Bissacco, A., Wu, B., Ng, A.Y.: Reading digits in natural images with unsupervised feature learning. In: NIPS workshop on deep learning and unsupervised feature learning. vol. 2011, p. 5 (2011) 7
25. Russakovsky, O., Deng, J., Su, H., Krause, J., Satheesh, S., Ma, S., Huang, Z., Karpathy, A., Khosla, A., Bernstein, M., et al.: Imagenet large scale visual recognition challenge. International Journal of Computer Vision **115**(3), 211–252 (2015) 7
26. Siam, M., Valipour, S., Jagersand, M., Ray, N.: Convolutional gated recurrent networks for video segmentation. arXiv preprint arXiv:1611.05435 (2016) 3
27. Srivastava, R.K., Greff, K., Schmidhuber, J.: Highway networks. ICML workshop on deep learning (2015) 3, 5
28. Szegedy, C., Vanhoucke, V., Ioffe, S., Shlens, J., Wojna, Z.: Rethinking the inception architecture for computer vision. In: Proceedings of the IEEE Conference on Computer Vision and Pattern Recognition. pp. 2818–2826 (2016) 1
29. Teerapittayanon, S., McDanel, B., Kung, H.: Branchynet: Fast inference via early exiting from deep neural networks. In: 23rd International Conference on Pattern Recognition. pp. 2464–2469 (2016) 3
30. Vaswani, A., Shazeer, N., Parmar, N., Uszkoreit, J., Jones, L., Gomez, A.N., Kaiser, Ł., Polosukhin, I.: Attention is all you need. In: Advances in Neural Information Processing Systems. pp. 6000–6010 (2017) 2
31. Vinyals, O., Fortunato, M., Jaitly, N.: Pointer networks. In: Advances in Neural Information Processing Systems. pp. 2692–2700 (2015) 2
32. Viola, P., Jones, M.: Rapid object detection using a boosted cascade of simple features. In: Computer Vision and Pattern Recognition, 2001. CVPR 2001. Proceedings of the 2001 IEEE Computer Society Conference on. vol. 1, pp. I–I. IEEE (2001) 3
33. Wan, L., Zeiler, M., Zhang, S., Cun, Y.L., Fergus, R.: Regularization of neural networks using dropconnect. In: Proceedings of the 30th International Conference on Machine Learning. pp. 1058–1066 (2013) 7
34. Williams, R.J.: Simple statistical gradient-following algorithms for connectionist reinforcement learning. Machine learning **8**(3-4), 229–256 (1992) 2, 6

# Appendix

In this supplementary material, we provide additional results on the CIFAR-10, CIFAR-100 and SVHN datasets to further demonstrate how SkipNet leverages different computational cost and accuracy to meet various computation and accuracy requirements. Moreover, we provide extra visualization to show the skipping ratio of different blocks of SkipNet to reveal the skipping patterns.

## 6  Trade-off Computational Cost and Accuracy

As described in the Eq. 5 of the main paper (we put the equation in Eq. 8 here for easy reference), our hybrid objective function leverages both the prediction error and the skipping reward. $\alpha$ is the tuning parameter that can balance the two components.

$$\min \mathcal{J}(\theta) = \min \mathbb{E}_{\mathbf{x}} \mathbb{E}_{\mathbf{g}} L_\theta(\mathbf{g}, \mathbf{x})$$
$$= \min \mathbb{E}_{\mathbf{x}} \mathbb{E}_{\mathbf{g}} \left[ \mathcal{L}(\hat{y}(\mathbf{x}, F_\theta, \mathbf{g}), y) - \frac{\alpha}{N} \sum_{i=1}^{N} R_i \right], \quad (8)$$

In conjunction with the hybrid objective function, we also discussed several gate designs of feed-forward gates (FFGate) and recurrent gates (RNNGate) with different level of parameter sharing and various computation overhead. Fig. 11 shows the gate designs used in this work. In our experiments, we use the computationally less expensive `FFGate-II` for deeper networks (greater than 100 layers) in our experiments and `FFGate-I` for shallower networks.

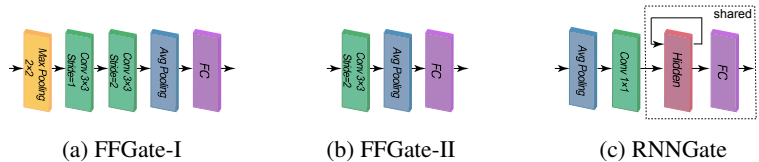

(a) FFGate-I    (b) FFGate-II    (c) RNNGate

Fig. 11: Gate designs used in the paper. (a) FFGate-I mainly consists of two 3x3 convolutional layers and has roughly 19% of the computation of the residual block. (b) FFGate-II is composed of one convolutional layer with stride of 2 and has about 12.5% of the computation of residual blocks. (c) RNNGate contains a one-layer LSTM with both input and hidden unit size of 10. It only has 0.04% of the computation of residual blocks and thus its computation cost is negligible

With larger $\alpha$, the model is trained to skip more layers leading to a more computationally light model and vice versa. We show the trade-off between number of skipped layers and the accuracy, with both feed-forward gates (FFGate) and recurrent gates (RNNGate) on the CIFAR-10, CIFAR-100 and SVHN datasets in Fig. 12.

We observe that for small $\alpha \leq 1.0$, the trade-off curve is relatively flat indicating that a large decrease in computation is accompanied by a negligible decrease in accuracy.

However, for larger $\alpha > 1.0$, both computation and accuracy decrease rapidly. By adjusting $\alpha$, one can trade-off computation and accuracy to meet various computation or accuracy requirements. We also note that though SkipNet with FFGate can skip similar number of layers as SkipNet with RNNGate, due to the computation overhead of FFGate, RNNGate is preferable.

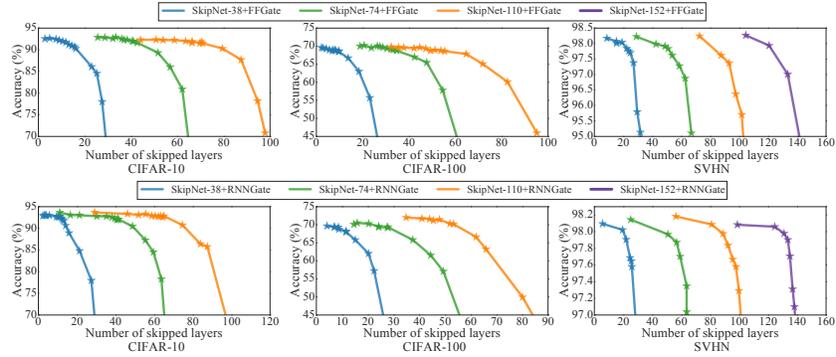

Fig. 12: Trade-off between computational cost and accuracy. More computation leads to better accuracy. With a small $\alpha$, the accuracy could remain roughly the same with significant drop in cost. With a large $\alpha$, both accuracy and computation drop substantially. One can find a critical point in the curve which well balances computation and accuracy. Though both gate designs can skip similar number of layers, the recurrent gate design is more computationally efficient due to the extremely low computational cost from the gate itself.

## 7 Skip Ratio of Different Blocks

In addition to the visualization of SkipNet skip ratio on CIFAR-10 provided in the main paper, we show the visualization of SkipNet on ImageNet (Fig. 13) and on CIFAR-100 (Fig. 14). The models augmented with gates can match the accuracy of the original ones. The skip ratio per block measures the ratio of images that skip the block. As shown in Fig. 13, SkipNet skips less blocks in the Group 1, 2 and 4 while skipping more blocks in Group 3. This indicates that those in Group 1, 2 and 4 are critical blocks for all inputs to extract accurate features. In addition, Group 1, 2 and 4 only have 3 or 4 blocks which lead to much less room for skipping compared to Group 2 with 23 blocks.

Fig. 14 shows the skip ratio of SkipNet-110 on CIFAR-100. Compared to SkipNet-101 on ImageNet, a more challenging dataset than CIFAR-100, SkipNet-101 on CIFAR-100 can reduce more computation while preserving the full network accuracy. While SkipNet skips intensively in almost all the groups, Group 2 skips less than the other two groups. This observation matches what we show of SkipNet-74 on CIFAR-10 in Fig.10a of the main paper, suggesting that in the residual network design, Group 2 is more critical on feature extraction than the other two groups on CIFAR-10.

These figures reveal the dynamic skipping patterns of SkipNet which also show that SkipNet can successfully identify critical blocks of a given input and skip unnecessary blocks to achieve a reduction in computation.

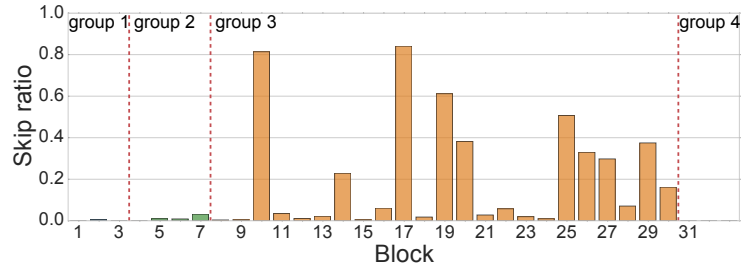

Fig. 13: Skip ratio per block of SkipNet-101 on ImageNet. SkipNet skips less blocks in the Group 1, 2 and 4 while skipping more blocks in Group 3 indicating that blocks in Group 1, 2 and 4 are critical on all inputs for feature extraction. Considering the number of blocks in Group 1, 2 and 4 (3 or 4 blocks), there is less room for skipping in those groups compared to Group 2 with 23 blocks

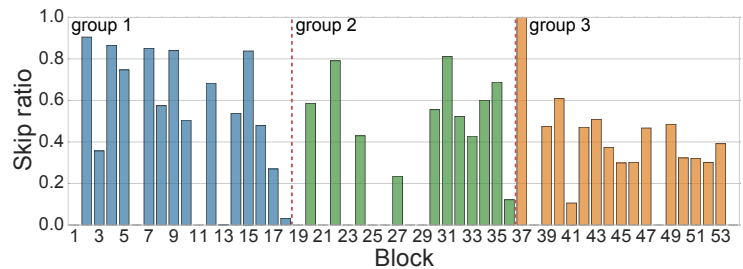

Fig. 14: Skip ratio per block of SkipNet-110 on CIFAR-100. While SkipNet skips intensively in almost all the groups, Group 2 skips less than the other two groups indicating that blocks in Group 2 are relatively more critical for the feature extraction on the CIFAR datasets